# Mimicking the Mavens: Agent-based Opinion Synthesis and Emotion Prediction for Social Media Influencers

Qinglan Wei[a], Ruiqi Xue[a], Yutian Wang[b], Hongjiang Xiao[b,*], Yuhao Wang[b], Xiaoyan Duan[a]
[a] The School of Data Science and Intelligent Media, Communication University of China, Beijing, 100024, China
[b] State Key Laboratory of Media Integration and Communication, Communication University of China, Beijing, 100024, China

## ABSTRACT

Predicting influencers' views and public sentiment on social media is crucial for anticipating societal trends and guiding strategic responses. This study introduces a novel computational framework to predict opinion leaders' perspectives and the emotive reactions of the populace, addressing the inherent challenges posed by the unstructured, context-sensitive, and heterogeneous nature of online communication. Our research introduces an innovative module that starts with the automatic 5W1H (Where, Who, When, What, Why, and How) questions formulation engine, tailored to emerging news stories and trending topics. We then build a total of 60 anonymous opinion leader agents in six domains and realize the views generation based on an enhanced large language model (LLM) coupled with retrieval-augmented generation (RAG). Subsequently, we synthesize the potential views of opinion leaders and predicted the emotional responses to different events. The efficacy of our automated 5W1H module is corroborated by an average GPT-4 score of 8.83/10, indicative of high fidelity. The influencer agents exhibit a consistent performance, achieving an average GPT-4 rating of 6.85/10 across evaluative metrics. Utilizing the "Russia-Ukraine War" as a case study, our methodology accurately foresees key influencers' perspectives and aligns emotional predictions with real-world sentiment trends in various domains.

## 1 INTRODUCTION

Opinion leaders, introduced by Paul Lazarsfeld and Elihu Katz in 1955, are highly regarded due to their influential status, expertise, or social prestige, functioning significantly in information distribution [1]. The rise of social media has spotlighted users such as public intellectuals and bloggers who hold sway over a vast audience; their ability to widely disseminate views and information empowers them as modern opinion leaders [2]. They mold public attitudes and behaviors, particularly utilising emotionally charged content for dramatic effects on social platforms [3, 4 ,5, 6]. Concrete examples include environmentally dedicated advocates promoting a circular economy, whose positive sentiments spur similar actions among followers. Moreover, advice from opinion leaders, like encouraging vaccination, has been instrumental in driving public behavior change [7]. Therefore, foreseeing opinion leaders' sentiments and viewpoints is vital in steering public opinions [8], allowing insights into impending societal movements and concentration points, thereby offering a futuristic perspective for policy framing, market surveying, and social dynamics analysis.

There is a significant change in the field of sentiment analysis, with a growing need for predictive analysis rather than reactive sentiment assessment in practice. Despite advances in the emotional computing of comments on social media, challenges remain due to the unstructured nature of views and their sensitivity to various contextual factors. These complexities prevent accurate predictions of the opinions and emotions of influencers on social platforms. Current approaches, such as Bidirectional Encoder Representation from Transformers (BERT) [9] for review understanding and aspect-based sentiment recognition, have made great strides in postmortem analysis, but have fallen short in pre-emptive insights. The predictive potential of sentiment analysis, especially in identifying the emotions of opinion leaders, remains an under-explored area.

In response to the above issues, the paper proposes a method (Multi-domain opinion leader agents emotion prediction, MOAEP) that uses LLM to simulate the viewpoints of opinion leaders in multiple domains and to achieve emotional predictions beforehand. This method consists of three modules: automatic question generation (AQG), multi-domain opinion leader agents construction (MOA), and pre-event prediction.

Firstly, in order to enable the agent to generate a more comprehensive response to a topic, the AQG module is formed by chain-based large language model (LLM) and vector dimension similarity retrieval. This realizes the generation of a finely selected question set in the 5W1H (Where, Who, When, What, Why, and How) format from a single topic. Furthermore, a knowledge based comprising 60 agent roles has been constructed. MOA is then built via the LLM chains and Retrieval-Augmented Generation (RAG) [10] framework, to receive the question set and simulate responses. Ultimately, through the simulated responses of multiple-domain opinion leader agents towards certain event topics, the opinions are extracted and clustered, and the emotions of the responses of each domain role are calculated, achieving pre-event prediction.

**Overall, the main contributions of the paper are threefold:**
- **Inquisitive AQG module:** It uses a single topic to auto-generate a diverse array of "5W1H" format questions,

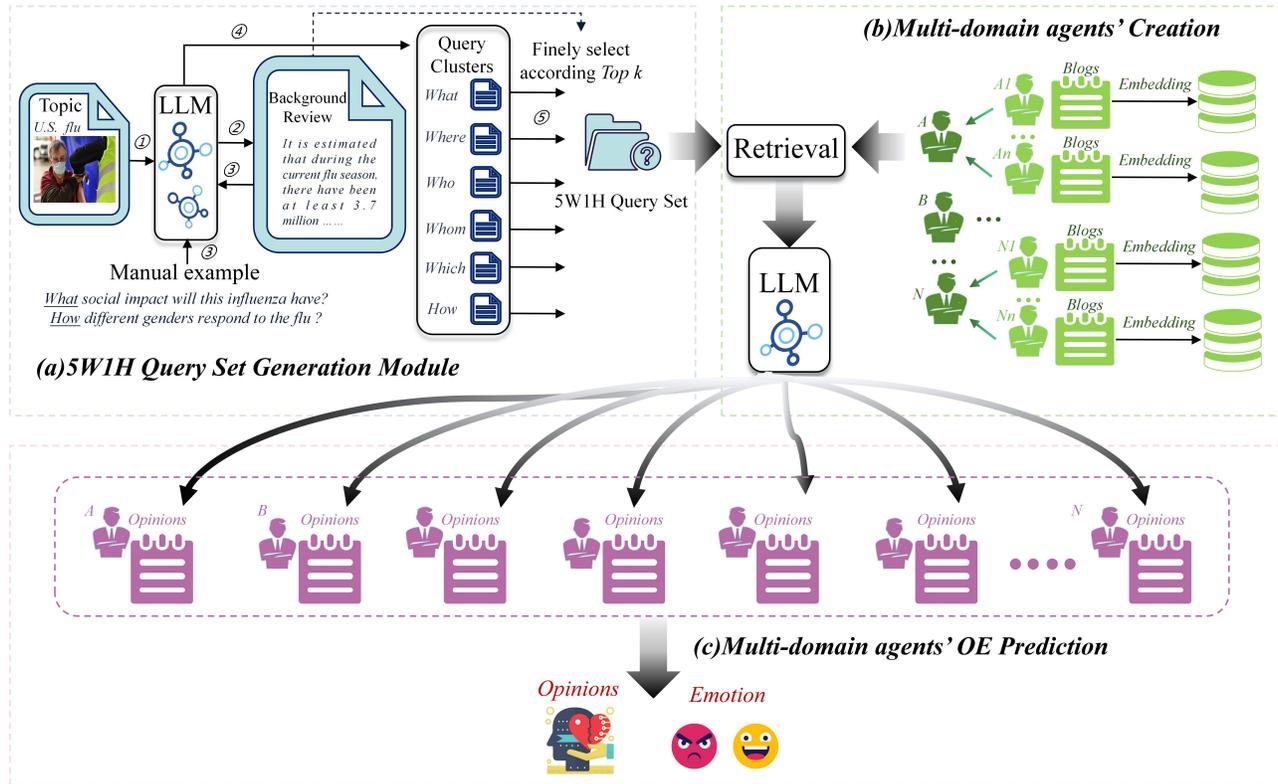

**Figure 1: The Pipeline of the MOAEP Method. The input is the name of the event for which the emotion needs to be predicted. Through module (a), the question set is automatically generated. Through module (b), the viewpoints of multi-domain opinion leader agents are generated. Through module (c), the response is analyzed to predict opinions and emotions.**

enabling thorough questioning of an agent on any specific topic. **The automatically generated question sets are excellently rated by GPT-4 for their topic relevance and format accuracy.**
- **Extensive MOA module:** It constructs 60 opinion leader agents across six key domains. These opinion leader agents are proved to **provide a satisfactory level of imitation of real-world opinion leaders upon evaluation through GPT-4.**
- **Prophetic MOAEP framework:** a pre-event prediction model, which amalgamates agent simulation and emotion calculation is successfully implemented in the "Russo-Ukrainian War" scenario. T**he emotional prediction results are found to be coherent with real-world responses, which validates its effectiveness.**

## 2 RELATED WORK

In the field of emotion analysis and emotion computing, there is a key shift from traditional emotion assessment to predictive analysis [11, 12, 13, 14]. The ability to predict emotions, especially for opinion leaders in social media, is increasingly important for understanding and effectively shaping public discourse [15, 16, 17]. Current research has made significant progress in using cutting-edge machine learning techniques to process structured data, such as movie reviews and social media feedback [11, 18]. For example, the BERT model is specially optimized to significantly improve the reading comprehension of review content [12, 19]; At the same time, deep learning techniques have also been developed to identify diverse emotional states from text, and these techniques have shown excellent results in cross-language and cross-platform emotion recognition and analysis [20, 21, 22].

However, the very nature of opinion data poses special challenges for sentiment analysis [23]. First, ideas are often presented in unstructured text, which increases the difficulty of extracting valuable information from the data [24, 25]. Second, views are influenced by a combination of factors such as time, environment, and individual differences, which complicates the construction of models that accurately predict changes in the mood [26]. Despite these challenges, researchers continue to explore the use of natural language processing techniques and dimensionality reduction algorithms to deeply analyze emotional expression in social media [27]. Some studies improve the analysis accuracy of blog text emotional tendencies by integrating vocabulary knowledge and text classification techniques[13, 28, 29]. Other studies have used models such as long short-term memory networks (LSTM) to gain a deeper understanding of the emotional information contained in the text [30]. The application of sentiment analysis has expanded beyond the understanding of individual emotions to predict and analyze broader social dynamics [31, 32]. During elections, for example, sentiment analysis is used to gauge the tendency of public opinion while identifying the spread of extremist sentiment in social media [15, 33]. The COVID-19 outbreak has also prompted researchers to use sentiment analysis techniques to extract public health-related information from social media texts [34].

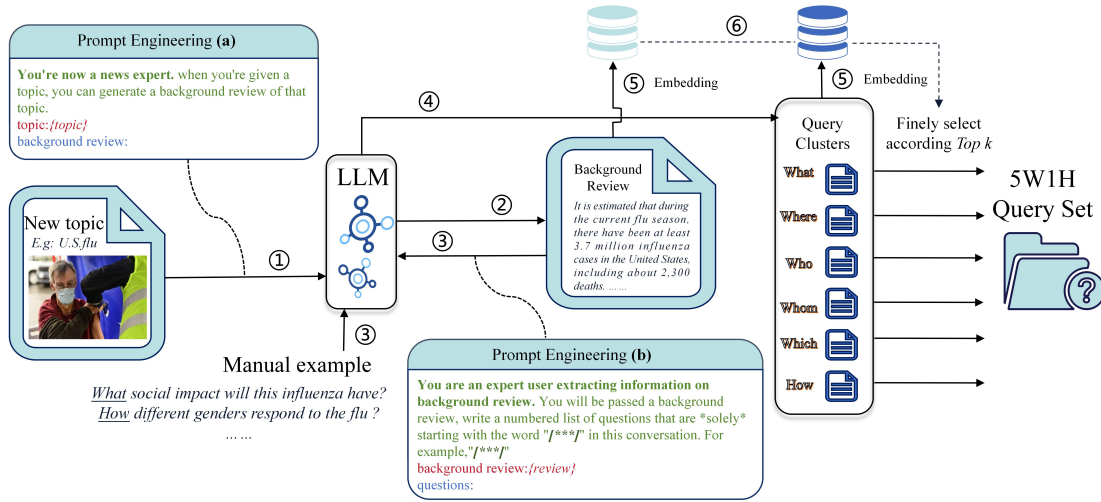

Figure 2: Detailed expansion of the Figure 1(a) module，AQG. Numbers 1 to 6 represent the steps in the module operation, and italicized characters represent content that can be modified during specific operations. (a)(b) are constructed prompt templates, where the green font represents the System, the red font represents the User, and the blue font represents the Output.

Although these research results have contributed to the development of the field, there are still significant shortcomings in predicting the emotions of social media opinion leaders. The varying social media content and the multiple factors affecting opinion formation urgently require researchers to develop more complex and accurate analysis models. These models need to be able to handle nonlinear relationships and complex interactions between textual data and their expressed emotions.

**To sum up, for practical application scenarios, the opinion and emotion prediction of opinion leaders on social media has become the core research focus. At present, there are few researches in this area, and the prediction accuracy is difficult to guarantee.**

## 3 METHOD

The overall construction process of the pre-event prediction system based on multi-domain opinion leader agents is shown in Figure 1. This system simulates the process from event occurrence to opinion expression by opinion leaders in different domains and can further predict their emotions.

Firstly, we construct a generation module from the topic event name to a curated set of questions in the 5W1H format, as shown in Figure 1(a). By leveraging the expansion, information extraction, and imitation capabilities of large models, we realize the automatic generation of serial selected question sets from a single word. To realistically mimic the viewpoints generated by opinion leaders in multiple domains towards an event, we establish a multi-domain opinion leader agent based on the RAG method [10], as shown in Figure 1(b). It relies on FAISS [35] for fast retrieval and generates viewpoint responses through LLM. Finally, the responses are input into the OE prediction module as shown in Figure 1(c). Further, we extract the opinions and predict the emotions of the viewpoints generated by the simulated opinion leaders. The details of each module above will be introduced in the following sections.

### 3.1 Automatic Question Set Generation (AQG)

To make the expression of opinions on a social event by opinion leader agents more comprehensive, we designed an automatic generation module from the topic to a curated set of questions (AQG). This module mainly consists of a chained LLM and similarity retrieval based on vector dimensions. The specific structure is shown in Figure 2. According to Matthew J. Salganik, the best starting point for social research is the "5 Ws": Who, What, Where, When, and Why [36]. Combining this with the six elements of current news and adding "how" to the "5 Ws", we set the question format to 5W1H, that is, questions beginning with "Who, What, Where, When, Why, and How". This module, characterized by robust generalizability, requires no initial data set. Furthermore, it showcases the flexibility, allowing for timely customization to meet transient requirements.

*3.1.1 Questions Generation.* Due to the limited generation capacity of LLM, it is impossible to generate a large number of questions directly from a topic in the form of a single word without using a dataset. Therefore, this paper employs Langchain to establish a chained LLM, calling LLM twice in succession with prompt engineering, making full use of its expansion, retrieval, and imitation capabilities.

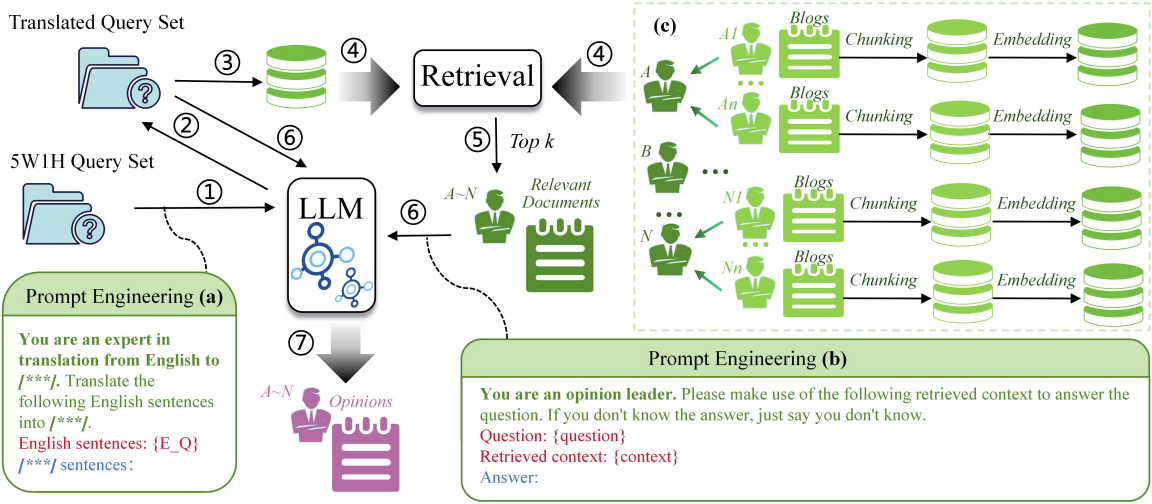

**Figure 3:** Detailed extension diagram of the Figure1(b) module，MOA. Numbers 1 to 7 indicate the steps in which this module operates. The green-dotted box shows the storage module process (c) for agent knowledge information. (a)(b) are the constructed prompt word templates, where the green font represents System, the red font represents User, and the blue font refers to Output.

Specifically, in step 1 of Figure 2, we construct a zero-sample prompt word template (a) using the classic three-stage prompt method. The System part adopts a role setting plus task description instruction method, the first sentence is the LLM role limited by this part (bold sentence), and the rest are specific task instructions, which maintain the characteristics of being brief and clear. System (a) sets the LLM task to expand the received topic name to generate the corresponding topic background. The User (a) {topic} pre-inputs "New Topic" reservation location. The format of the New topic is like the topic example given in the figure "U.S. flu", just a text format topic name is needed, without limitation on the length or number of words. Step 1 connects the prompt word template (a) constructed by Langchain and "New Topic" with LLM, and step 2 calls LLM for the first time to output the background text corresponding to the topic.

In step 3, a second keyword template (b) is constructed to extract question sets from the generated extended text. Template (b) also uses the classic three-part hint method, and system (b) also includes role-definition sentences and task instructions. In addition to this, *[\*\*\*]* is added to the system (b) for different format question hint examples, expanding keyword (b). This is due to the limitations of current LLM capabilities, and the number of questions generated at one time is limited. Assuming the threshold for generating questions is $\theta$, it is not possible to force the generator to produce m questions that exceed the number $\theta$ at one time, otherwise, there will be $(m-\theta)$ duplicate questions. The total number of different format questions generated at one time is limited by $\theta$. To generate as many questions as possible and format them correctly, *[\*\*\*]* expands the keyword template (b) into six similar format keyword templates, each responsible for extracting What, Where, Who, Whom, Which, and How Questions, thereby making the threshold $(6*\theta)$, expanding the capability of LLM to output the number of questions. The first *[\*\*\*]* in the keyword template (b) is input from one of the 5W1H words, and the second *[\*\*\*]* corresponds to sentence examples that begin with the word in a manual case. Manual examples are not complicated, only 6 sentence examples are needed, each sentence starting with a 5W1H word. Since this example is only to let LLM understand the 5W1H format questions, the specific content is not the most important, and there is no need for complicated and excessive manual examples, just set it once to apply to each different topic input. The same as User(a), User(b) reserves the position for the pre-input "Background review" at the {review}. In step 4, LLM is called to output the corresponding initial 5W1H question set $Q$:

$$Q\{Q_i\{q_j\}\}, (i = 1 \sim 6, j = 1 \sim \theta), \quad (1)$$

The above method realizes the automatic generation of a large number of question sets in a data set-free state.

**3.1.2 Questions Selection.** Next, the initial question set $Q$ is further refined. After six types of question sets $Q_i$ are compared with the "Background review" for similarity, each question set selects the top $k$ questions for similarity, summarizes, and outputs the refined question set $Q'$. First, in step 5, the release model "text2vec-base-multilingual" [37] is used to transform the Background review and $Q_i$ into a vector format. After the "text2vec-base-multilingual" model encodes the sentences, each sentence will be mapped to a 384-dimensional dense vector space. At this point, the vector $B$ of the Background review and the vectorized six types of question sets $Q_i'$ are obtained:

$$B = [v_1, v_2, \cdots, v_x], (x = 384), \quad (2)$$

$$Q_i' = \{q_j'\}, q_j' = [v_1, v_2, \cdots, v_x], (i = 1 \sim 6; j = 1 \sim \theta, x = 384), \quad (3)$$

At the same time, an L2 distance index is established for each vector $q_j'$ in $Q_i'$ through FAISS, which is conducive to the rapid calculation of sentence similarity. In step 6, the L2 distance $D_i(q_j', B)$ between each vector $q_j'$ in $Q_i'$ and vector $B$ is calculated, and the $j$ label of $k$ $q_j'$ vectors closest to vector $B$ is returned. The $j$ label corresponds to $q_j$ in formula 1 and returns the original question text. The formula for calculating $D_i(q_j', B)$ is as follows:

$$D_i(q_j', B) = \sum_{v=1}^{x}(q_j\_v - B\_v)^{\wedge}2, (x = 384), \quad (4)$$

Due to the randomness of LLM output, the value of $\theta$ is not fixed, but as long as the $k$ set is less than $\theta$, the above method always holds. The selected question set returned from the six types of question sets $Q_i$ is summarized to get the final $Q'$.

## 3.2 Multi-domain Opinion Leader Agents (MOA)

To predict opinion leaders' emotions in advance at the cost of the smallest resources and time consumption, we construct the MOA, using the LLM and RAG frameworks. Ethical risks are avoided while simulating various opinions that opinion leaders in multiple domains would express after an event occurs, as shown specifically in Figure 3. This module mainly consists of storage for multi-agent knowledge information, quick search, and the LLM, aiming to respond to multiple agents simultaneously.

*3.2.1 Agents Knowledge Information Storage.* This module displays the process of building a vectorized information library for generalized roles of opinion leaders across multiple domains, as shown in Figure 3(c). In this module, to avoid the ethical risks brought by obtaining and displaying entity role information, we first propose a concept of generalized role. This generalized role consists of multiple anonymous entity roles with similar views or domains, and the information of multiple entity roles is bound and output, thereby reducing the risk of being recognized after the entity role is anonymized. In this module, we construct generalized roles of opinion leaders in $N$ domains, such as $A \sim N$ in Figure 3(c). The generalized role of each domain consists of n entity opinion leader roles, such as $A_1 \cdots A_n \sim N_1 \cdots N_n$ in Figure 3(c).

Specifically, looking at the storage construction process in Figure 3(c), the initial knowledge information of each opinion leader entity role is a text-format blog, which is a long text composed of viewpoints generated by the entity role on a large number of events. Subsequently, each blog is treated in chunks, dividing the long text into $m$ fixed-length small paragraph texts $C$ for storage. After establishing a corresponding anonymous vector library $A_i$ for each entity role, we encode the chunked text data $C$ corresponding to each role, ultimately transforming it into vector format storage:

$$A_i = \{C_j\}, C_j = [v_1, v_2, \cdots, v_x], (i = 1 \sim n, j = 1 \sim m, x = 384) \quad (5)$$

*3.2.2 Agents Construction Process.* The construction of the multi-domain opinion leader agents mainly relies on the RAG framework, with LLM serving as the brain of the Agent, and retrieval serving as the opinion imitation tool for each role, with the specific process outlined in steps 1~7 of Figure 3. Subject to 5W1H format restrictions, the automatically generated question set is in English, while in (c), different language data will appear due to the different regions of the opinion leaders. To align the languages, in step 1, a classical three-stage method is used to establish prompt word templates, which are used to translate English questions into questions in a specific language. The System still adopts the method of adding role description plus task description, and {E_Q} in the User reserves a position for the question to be translated. The three placeholders *[\*\*\*]* in System and Output are filled with the same language according to the data language in (c). Similarly, Langchain is used to connect the 5W1H English question set and prompt template (a) with LLM, and in step 2, LLM is called to translate the question set into a specific language. In step 3, each question is converted into a vector form $q$:

$$q = [v_1, v_2, \cdots, v_x], (x = 384), \quad (6)$$

In step 4, the question vector $q$ and entity role vector data $A_i$ in (c) are rapidly retrieved for similarity. The retrieval uses the L2 distance between vectors:

$$D_i(C_j', q) = \sum_{v=1}^{x}(C_j\_v - q\_v)^{\wedge}2, (x = 384), \quad (7)$$

Step 5 returns the top $k$ text segments in (c) that are similar to the question. In step 6, the classical three-segment method is still used to construct the core prompt template of the agents. The System prescribes the role and task of the agents. User includes both the question {question} and the retrieved text {context}. Langchain reconnects these parts with LLM, simulating the generation of opinion leader views in step 7. Note that when the opinion leader's generalized role is activated, its corresponding opinion leader entity roles are then awakened, but the awakening process of the opinion leader entity roles is a hidden state and the outside world cannot see the specific responses of each entity role. Only after the responses of all opinion leader entity roles are completed, these responses are packaged as a whole and outputted to the corresponding opinion leader's generalized role.

## 3.3 Pre-event Prediction

After obtaining the responses generated by the opinion leader agents in various fields to the questions, we further analyze and mine the responses using existing methods. We explore the main opinions of opinion leaders in each field regarding the event and make predictions about their emotions.

*3.3.1 Viewpoint Extraction and Clustering Analysis.* To explore the primary viewpoints contained in the responses, we employ the existing open-source method, OpinionExtraction [38], for viewpoint extraction. We then use k-means clustering on the extracted viewpoints to grasp the main categories of viewpoints. Specifically, we first divide the viewpoints generated in response in to sentences, setting obvious skip words unrelated to viewpoints, such as "I don't know, number, system prompt," and remove sentences containing these words. We cycle through the cleaned sentences and use the dependency-based viewpoint extraction method in OpinionExtraction to extract the main viewpoints of each sentence and generate new sentences. After segmenting the opinion sentences, we vectorize the sentences using the TF-IDF method [39], where D in IDF is the summary document of opinion sentences. Next, we use KneeLocator to automatically find the optimal number of clusters for K-means. Then we cluster the vectorized opinion sentences and conduct visual analysis.

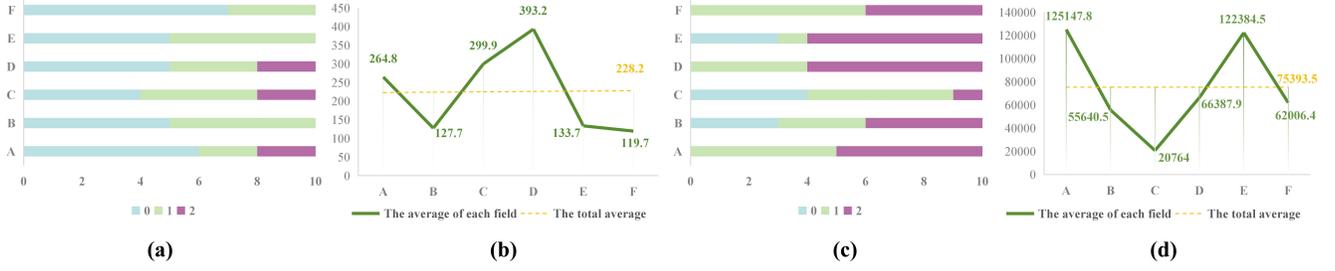

Figure 4: Dataset statistics. (a) shows the distribution of Weibo followers for each agent role in each field, (b) shows the average number of Weibo followers for each agent role in each field, (c) shows the distribution of text lengths for the crawls of every agent role in each field, and (d) shows the average text length for the crawls of every agent role in each field. Throughout these four figures, ABCDEF respectively represents the fields of politics, economics, technology, society, entertainment, and military. In (a), label 0 represents roles having less than or equal to 1 million followers, label 1 represents roles with more than 1 million but less than or equal to 5 million followers, and label 2 represents roles with over 5 million followers. In (b), the unit for the y-axis is ten thousand. In (c), label 0 represents roles with a text length of fewer than 10,000 words, label 1 represents roles with a text length greater than 10,000 but less than or equal to 50,000 words, and label 2 represents roles with a text length of over 50,000 words. In (d), the unit for the y-axis is one.

*3.3.2 Calculation and Analysis of Emotion.* For sentiment analysis of the response content, we mainly use the existing algorithm for the automatic construction of domain-adaptive sentiment dictionaries based on semantic rules [40]. The sentiment calculation for each response content is done through the automatically constructed sentiment dictionaries. First, all response texts are input as a corpus. According to the corpus adaptation threshold formula proposed in the algorithm for automatic construction of domain-adaptive sentiment dictionaries based on semantic rules, new words are discovered and emotionless new words are screened out based on their emotional tendencies. This information is combined with the fixed emotion dictionary to construct a sentiment dictionary exclusive to the response text. According to the constructed sentiment dictionary, the emotional value of each response text is calculated. The sentiment score is a continuous number from -1 to 1, and the calculation results for the emotions within the generalized roles of the opinion leaders in each field are summarized, and statistically analyzed.

## 4 EXPERIMENTS

### 4.1 Dataset

In the experiment, a self-built dataset is used for constructing the knowledge base of multi-domain opinion leader agents. We define six generalized opinion leader roles in the fields of economics, military, technology, society, entertainment, and politics. Each composes of 10 agent opinion leader roles. The data source comes from Weibo, which is a popular social media in China for sharing and discussing personal information, life activities, and celebrity news [41]. The selected agent roles that serve as opinion leaders all have more than 100,000 followers on the Weibo platform. The text content posted by these users on Weibo over the past month is gathered and preserved in Word format as the original document. Specific statistics about the dataset can be seen in Figure 4. The average number of Weibo followers for the 60 roles is as high as 2.282 million, with an average of 75,000 words of text content retrieved for each role.

### 4.2 Implementation Details

As the knowledge base text is in Chinese, the LLM uses the open-source ChatGLM3-6B [42] model. The max_length is set to 8192, top_p is set to 0.8 and temperature is set to 0.7. In all experiments, top k is set to 2. All experiments are conducted on high-performance computing platform with adequate GPUs.

### 4.3 Evaluation

*4.3.1 AQG.* To test the performance of the AQG module, we select topics from three major categories: health, economics, and psychology. We choose 5 topics from each category, creating a total of 15 topics to input into the module for generation evaluation.

*4.3.1.1 Evaluation Method.* Three-dimensional indicators are set to evaluate the AQG module. The first dimension evaluates the effect of the "background review" generated in the intermediate

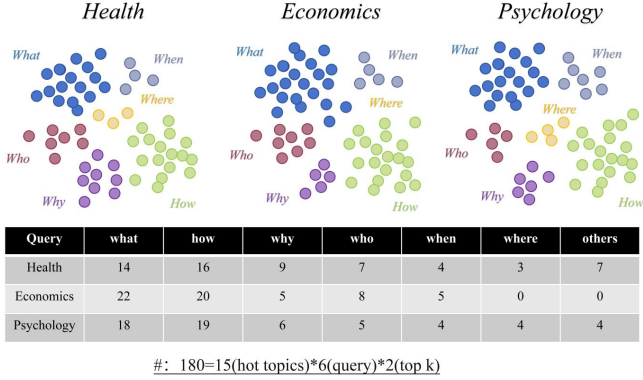

| Query | what | how | why | who | when | where | others |
|---|---|---|---|---|---|---|---|
| Health | 14 | 16 | 9 | 7 | 4 | 3 | 7 |
| Economics | 22 | 20 | 5 | 8 | 5 | 0 | 0 |
| Psychology | 18 | 19 | 6 | 5 | 4 | 4 | 4 |

#: 180=15(hot topics)*6(query)*2(top k)

**Figure 5: The number and format distribution of selected question sets.** This figure presents the distribution of questions in different formats.

**Table 1: Evaluate AQG by GPT**

| - | RBT | RQQ | RBQ |
|---|---|---|---|
| Health | 10.000 | 9.942 | 9.000 |
| Economics | 10.000 | 10.000 | 8.175 |
| Psychology | 10.000 | 9.967 | 7.983 |
| **Avg.** | **10.000** | **9.969** | **8.386** |

**Table 2: Manual evaluation of AQG**

| - | RBT | RQQ | RBQ |
|---|---|---|---|
| Health | 8.067 (2.813) | 9.622 (0.135) | 8.289 (0.810) |
| Economics | 8.667 (1.973) | 8.667 (5.333) | 8.933 (3.413) |
| Psychology | 9.467 (0.373) | 9.844 (0.001) | 9.133 (0.618) |
| **Avg.** | **8.733 (0.493)** | **9.378 (0.392)** | **8.785 (0.195)** |

link of the module, which is the "Relevance Between Topic and News Background (RBT)". The other two dimensions are used to evaluate the generated questions, which are "Whether the Question Belongs to 5W1H and its Relevance to 5W1H (RQQ)" and "Relevance Between Question and News Background (RBQ)". Both GPT4 and manual evaluation methods are used to score the module's generation results on a scale of 1-10 across these three dimensions - the higher the RBT score, the greater the relevance between news background and theme; the higher the RQQ score, the closer the format of the question to the 5W1H format; the higher the RBQ score, the higher the relevance between news background and question.

In terms of GPT evaluation, GPT4 is used for testing. In manual evaluation, a quarter of the original questions from the three major categories are randomly selected for evaluation. After a simple training for evaluators on the understanding of evaluation dimension meanings, each question is scored anonymously by three evaluators. The final score for each question is the average of the three evaluators' scores, and its variance is also calculated.

*4.3.1.2 results.* The module successfully generates 180 selected questions from 15 topics, with the number and format distribution as shown in Figure 5. For different topics, the question sets generated contain more "what" (depicted by blue solid circles) and "how" (depicted by green solid circles) queries, while "where" (depicted by yellow solid circles) queries are the least. This phenomenon occurs because the Language Model does not always strictly follow the prompt template during generation, resulting in a small portion of questions in other formats.

Therefore, when the top 2 similar questions are selected, questions in other formats may be extracted. Consequently, the final quantity of questions in each format is not entirely equal, but essentially, questions in all formats are generated.

The evaluation results of the experiment in three dimensions by GPT and human evaluation are shown in Tables 1 and 2, respectively. As can be seen from the GPT results, the module performs exceptionally well on all three dimensions, with the best two dimensions being RBT and RQQ. This demonstrates that the news background generated in the intermediate step is highly relevant to the original input topic, and the automatically generated questions also closely conform to the preset 5W1H format. In the manual evaluation, RBT, RQQ, and RBQ all still perform quite well and have lower variances. The best-performing dimension is RQQ, while the average scores for the other two dimensions are also close to 9, indicating that the selected questions are also quite relevant to the news background. **The overall result indicates that the question sets produced through automation receive high praise from GPT-4 for their pertinence to the topic and precise formatting.**

*4.3.2 MOA.* The opinion leaders' entity roles are extracted. Based on the total number (60), sampling is conducted using a stratified (by field) sampling method. According to the 1:5 sampling principle, 2 agent roles are randomly selected from each layer, totaling 12 agents. An evaluation test is then conducted by simulating 10 questions.

*4.3.2.1 Evaluation Method.* Opinion leader agents belong to a type of role-based agent. Therefore, the evaluation indicators adopt the three dimensions set up in RoleLLM [43], namely, "Response Accuracy to Commands (RAW)", "Ability to Imitate Speech Styles Associated with Specific Roles (CUS)" and "Role-Specific Knowledge and Memory of the Model (SPE)". Each dimension is scored on a scale of 1-10. GPT4 is used to rate the responses to ten questions from the 12 agents based on these three evaluation dimensions. The final score for each agent in a certain dimension is the average score of the ten evaluation questions. The higher the RAW score, the more accurately the agent responds to commands. The higher the CUS score, the better ability the model has at imitating the speech style associated with a particular role. The higher the SPE score, the better the agent utilizes role-specific knowledge.

*4.3.2.2 results.* The three-dimensional evaluation results of the generalized role agents in each field are shown in Table 3. The model scores relatively high in the CUS and SPE dimensions, with average indicators of 7.558 for CUS and 8.175 for SPE. This indicates that the model performs well in imitating role language style and utilizing role-specific knowledge. However, some field models score relatively low in the RAW dimension. This is because when the question set does not belong to the domain knowledge of the model, the corresponding questions cannot be answered, resulting in a RAW score of 0. However, this performance is normal. For example, when an entertainment role is asked to answer financial questions, the RAW score is low.

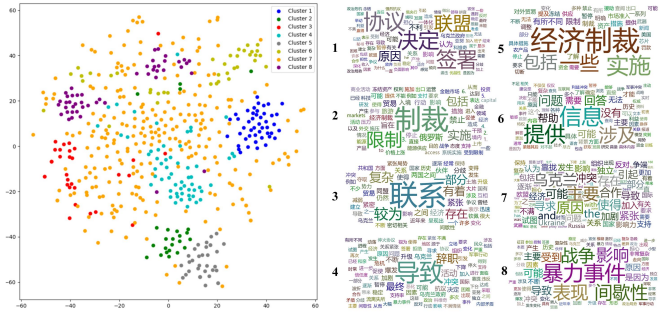

Figure 6: The clustering results after opinion extraction and the display of word clouds for each category, with numbers 1-8 in the right diagram respectively representing the categories in the left diagram.

Table 3: Agents evaluation in three dimensions

| Domain | RAW | CUS | SPE | Avg. |
|---|---|---|---|---|
| Politics | 6.250 | 7.800 | 7.00 | **7.017** |
| Economics | 4.200 | 7.700 | 6.900 | **6.267** |
| Technology | 3.750 | 7.950 | 8.350 | **6.683** |
| Society | 5.950 | 7.700 | 8.750 | **7.467** |
| Entertainment | 3.100 | 8.150 | 9.250 | **6.833** |
| Military | 5.600 | 6.050 | 8.800 | **6.817** |
| **Avg.** | **4.808** | **7.558** | **8.175** | **6.847** |

Table 4: Emotional calculation results in various domains

| Domain | Avg. |
|---|---|
| Politics | -0.0538 |
| Economics | -0.0588 |
| Technology | -0.0531 |
| Society | -0.0635 |
| Entertainment | -0.0747 |
| Military | -0.0909 |

Since the answer does not exhibit model hallucination and aligns with the content of the model's knowledge base, the SPE performance is high. **Overall, the model's three-dimensional evaluation results average a score of 6.847, indicating that the agent can, to a certain degree, impersonate opinion leaders in a real-world environment.**

*4.3.3 Pre-event Prediction.* Using "Russo-Ukrainian War" as an input topic example, the Figure 1 (a) (b) modules successfully simulate the response content of 60 roles to this topic, with a runtime of 840.54s. Next, an analysis of opinion extraction, clustering, and sentiment calculation results are conducted.

*4.3.3.1 Viewpoint Extraction and Clustering Analysis.* After sentence segmentation and cleaning, 550 opinion sentences are successfully extracted from all the response viewpoints. After the clustering module, the 550 opinion sentences are clustered into eight major categories of viewpoints. The scatter plot of the clustering results and the word cloud images for each category are shown in Figure 6. Most classes show a tight scatter distribution and clear clustering. The visualized word cloud images for each category display the specific opinion content of each class. For instance, in Category 3, the word cloud reflects the opinion leaders' belief in the complex and close relationship between Russia and Ukraine. In category 8, the word cloud depicts intermittent violent incidents between Russia and Ukraine. **By simulating and extracting the viewpoints of opinion leaders, we can largely understand and predict their views on the events.**

*4.3.3.2 Calculation and Analysis of Emotion.* Finally, we conduct sentiment calculations on the response texts of the generalized role agents in each field. The results are summarized in Table 4. As can be seen from the table, the opinion sentiment of the generalized role of the six major field opinion leaders toward the "Russia-Ukraine War" event is mildly negative. Among these, the political, economic, and technology fields show the lightest negative sentiments, the social and entertainment fields show moderately negative sentiments, and the military field has the heaviest negative sentiment. This predicted result also largely aligns with actual sentiment tendencies in each field about the "Russia-Ukraine War" event in real-world settings. **Therefore, using opinion leader role agents to emulate opinion responses can achieve sentiment prediction prior to real-world outcomes, from a single event word to the sentiment of opinion leaders in each field, and this prediction has a certain reference value.**

## 5  CONCLUSIONS

In this paper, we propose a MOAEP method based on large language model to simulate opinion leaders' emotions across multiple domains for pre-event emotional prediction. Firstly, we establish a Question Set Generation module (AQG) composed of an LLM chain and vector-similarity retrieval, transforming a single topic into a large selection of 5W1H-format curated questions. Secondly, we complete a Multi-domain Opinion Leader Agent module (MOA), which builds 60 agents across six major domains: politics, economics, science and technology, society, culture, and military. Here, we propose a concept of generalized roles to evade ethical risks. Ultimately, we extract opinions and calculate sentiments from agent responses. This process simulates the opinions generation and emotional analysis of multi-domain opinion leaders on a certain topic, which realizes a crucial paradigm shift from post-event analysis to pre-event prediction. According to quantitative and qualitative evaluation results, we can assert that this method enables pre-event prediction based on opinion leaders. In the future, adhering to our MOAEP framework, we can construct more different agents to simulate real-world roles, exploring their opinions, emotions, behaviors, etc., to achieve more elaborate pre-event prediction.

# APPENDIX

**Figure 1, Prompt for GPT4 evaluation in paper 4.3.1.1.**

| Relevance of background review and topic. | Whether it is a 5w1h question and its relevance to a 5w1h question. | The relevance of the question to the background review. |
|---|---|---|
| # Role: Text similarity detector<br><br>## Goals<br>Receive a topic and a background review. Detect the similarity between topic and background review, and give a similarity score.<br><br>## Constraints<br>Only assign scores within the range of 1 to 10.<br><br>Example:<br><br>How can patients manage the time and resources to eat healthy food in the hospital?<br><br>Your question closely aligns with the context of the background review, focusing on how patients can overcome the challenges faced when trying to eat healthily in a hospital. Considering its relevance to the subject at hand, I would give this question a similarity score of 9. It doesn't directly parallel a statement in the review but is closely tied to the overall theme. | # Role: 5W1H Query Analyst<br><br>## Goals<br>Receive an inquiry and delineate whether it falls under the 5W1H questions category. 5W1H questions commence with "What, Where, Who, When, Why, How". If the inquiry strictly follows the 5W1H format, promptly assign it a perfect score of 10. Conversely, if an inquiry does not adhere to the 5W1H structure, evaluate its correlation to the 5W1H query format and assign a score accordingly.<br><br>## Constraints<br>Only assign scores within the range of 1 to 10. | # Role: Question similarity detector<br><br>## Goals<br>Your primary task is to receive a question, then identify its relevance to a given context piece—referred to as the "background review". Next, you need to score the degree of similarity between the question and the background review.<br><br>## background review<br>{……}<br><br>## Constraints<br>Only assign scores within the range of 1 to 10. |

**Figure 2, Prompt for GPT4 evaluation in paper 4.3.2.1.**

| RAW | CUS | SPE | Similarity |
|---|---|---|---|
| ## Goal<br>You will receive text and events, analyze the text's evaluation of the event and the relevance of the event, and score based on the relevance.<br>## Event<br>{……}<br>## Constraints<br>Events are scored according to their relevance, with a maximum score of 10 and a minimum of 0. | ##Goal<br>You will receive a text, and compare the language style of the text with the text in your knowledge base. Score the similarity between the two language styles.<br>## Constraints<br>The score is based on the similarity of language style, with the highest score being 10 points and the lowest score being 0 points. The higher the score, the stronger the similarity in language style. | ##Goal<br>You will receive the text, please conduct a comprehensive comparative analysis of the text content and the knowledge base at the level of viewpoints and knowledge points, and calculate the score.<br>## Constraints<br>The highest score is 10 points and the lowest score is 0 points. The higher the score, the higher the similarity of opinions and knowledge points. | ##Goal<br>You will receive two pieces of text. Please calculate the similarity between the two pieces of text and rate them. ## Constraints<br>The higher the similarity between two pieces of text, the higher the score. The highest score is 10 points and the lowest score is 0 points. |

**Table 1: Topic input of MOA module in paper 4.3.1.1.**

| Health | Economics | Psychology |
|---|---|---|
| Winter flu epidemic. | China's stock market. | Interpersonal communication disorder. |
| Eating health food in the hospital. | China's new energy vehicle production and sales have been the world's first for nine consecutive years. | Marriage. |
| Zombie deer virus detected in the United States. | Microsoft acquires Activision Blizzard. | Anxiety. |
| Online medicine purchases in Shanghai can use medical insurance. | South Korea's real estate crisis. | Child psychiatry. |
| Sinovac's COVID-19 vaccine has ceased production. | The 30th APEC summit was held in San Francisco, USA. | Depression symptoms. |

**Table 2: In paper 4.3.2.1, the input problems are evaluated.**

| Question |
|---|
| How do you feel about the blockade of the Bab el-Mandeb Strait by the Houthis? |
| What are your views on the Palestinian-Israeli conflict? |
| What do you think of the war between Russia and Ukraine? |
| How do you feel about the decline in the stock market? |
| What do you think of the US presidential election? |
| How do you feel about Texas becoming independent? |
| How do you feel about Tesla's price increase? |
| How do you feel about the US Department of Commerce's claim to curb China's scientific and technological development? |
| What do you think of the popularity of new Chinese styles? |
| How do you feel about the decline of housing prices in China? |